\pdfoutput=1
\documentclass[11pt]{article}

\usepackage[preprint]{acl}

\usepackage{times}
\usepackage{latexsym}

\usepackage[T1]{fontenc}
\usepackage[utf8]{inputenc}

\usepackage{microtype}

\usepackage{inconsolata}

\usepackage{graphicx}

\usepackage{amsmath}
\usepackage{amssymb}
\usepackage{amsthm}

\usepackage{booktabs}
\usepackage{tabularx}
\usepackage{subfig}
\usepackage{multirow}
\usepackage{array,tabularx,booktabs,multirow}
\newcolumntype{C}{>{\centering\arraybackslash}X}
\usepackage{pifont}
\usepackage{xcolor}
\usepackage{enumitem}

\usepackage[most]{tcolorbox}
\tcbset{
  takeawaysstyle/.style={
    colback=orange!20,        % 背景色（橙色的20%透明度）
    colframe=orange!80!black, % 边框颜色（橙色加黑色调）
    fonttitle=\bfseries,      % 标题加粗
    coltitle=white,           % 标题颜色
    sharp corners,            % 直角边框
    boxrule=0.8pt,            % 边框线宽
    top=3pt, bottom=3pt, left=6pt, right=6pt, % 内边距
    boxed title style={
      colback=orange!80!black, % 调浅颜色
      sharp corners,
      boxrule=0pt,
      left=4pt, right=4pt, top=1pt, bottom=1pt,
    }
  }
}

\title{GloSS over Toxicity: Understanding and Mitigating Toxicity in LLMs via Global Toxic Subspace}

\author{
 \textbf{Zenghao Duan}\textsuperscript{1,2 *},
 \textbf{Zhiyi Yin}\textsuperscript{1 *},
 \textbf{Zhichao Shi}\textsuperscript{1,2 *},
 \textbf{Liang Pang}\textsuperscript{1 \(\dagger\)},
\\
 \textbf{Shaoling Jing}\textsuperscript{1},
 \textbf{Jiayi Wu}\textsuperscript{3},
 \textbf{Yu Yan}\textsuperscript{1,2},
 \textbf{Huawei Shen}\textsuperscript{1},
 \textbf{Xueqi Cheng}\textsuperscript{1},
\\
 \textsuperscript{1}Institute of Computing Technology, Chinese Academy of Sciences, Beijing, China
 \\
 \textsuperscript{2}University of Chinese Academy of Sciences, Beijing, China
 \\
 \textsuperscript{3}Dalian University of Technology, Liaoning, China
\\
 \small{
     \href{mailto:email@domain}{\{duanzenghao24s, yinzhiyi, pangliang, shenhuawei, cxq\}@ict.ac.cn}
 }
}

\begin{document}

\maketitle

% 处理特殊符号脚注
\stepcounter{footnote}
\footnotetext[\value{footnote}]{* Equal Contributions}
\stepcounter{footnote}
\footnotetext[\value{footnote}]{\(\dagger\) Corresponding author}

\begin{abstract}
This paper investigates the underlying mechanisms of toxicity generation in Large Language Models (LLMs) and proposes an effective detoxification approach.
Prior work typically considers the Feed-Forward Network (FFN) as the main source of toxicity, representing toxic regions as a set of toxic vectors or layer-wise subspaces.
However, our in-depth analysis reveals that the \textbf{global toxic subspace} offers a more effective and comprehensive representation of toxic region within the model.
Building on this insight, we propose \textbf{GloSS} (\textbf{\underline{Gl}}obal T\textbf{\underline{o}}xic \textbf{\underline{S}}ubspace \textbf{\underline{S}}uppression), a lightweight, four-stage method that mitigates toxicity by identifying and removing the global toxic subspace from the parameters of FFN.
Experiments across a range of LLMs show that GloSS achieves state-of-the-art detoxification performance while preserving the models’ general capabilities, without requiring large-scale data or model retraining. 
\textcolor{red}{WARNING: This paper contains context which is toxic in nature.} 
\end{abstract}

\section{Introduction}

Large language models (LLMs) have shown impressive capabilities in various domains~\cite{brown2020languagemodelsfewshotlearners,xin2024deepseekproverv15harnessingproofassistant,gu2025surveyllmasajudge}.
However, they also have risks of toxicity generation, which may lead to undesirable effect in real-world applications~\cite{ma2025safetyscalecomprehensivesurvey}.
To mitigate toxicity, tuning-based methods such as Supervised Safety Fine-Tuning (SSFT)~\cite{ouyang2022training} and Direct Preference Optimization (DPO)~\cite{dpo2023} have been widely adopted, improving LLM safety. However, aligned models can still be bypassed by crafted attack prompts~\cite{yan2025benignimporttoxicjailbreaking}.
As a result, recent researches have shifted toward analyzing the mechanisms of LLMs, aiming to understanding and locating the regions that elicit toxicity~\cite{suau2024whisperingexpertsneuralinterventions,wang-etal-2024-detoxifying}.

Toxic behaviors are often attributed to the Feed-Forward Network (FFN) of Transformer blocks, with two prevailing views proposed. One line of research, such as \citet{lee2024a}, identifies the toxic region as a set of toxic vectors, and argues that DPO mitigates toxic outputs by bypassing the region. Another approaches, exemplified by ProFS~\cite{uppaal2025model}, posit that toxicity resides in layer-wise toxic subspaces, identified via Singular Value Decomposition (SVD) of embedding differences between toxic and non-toxic prompts at each layer.

\begin{figure}
    \centering
    \includegraphics[width=\linewidth]{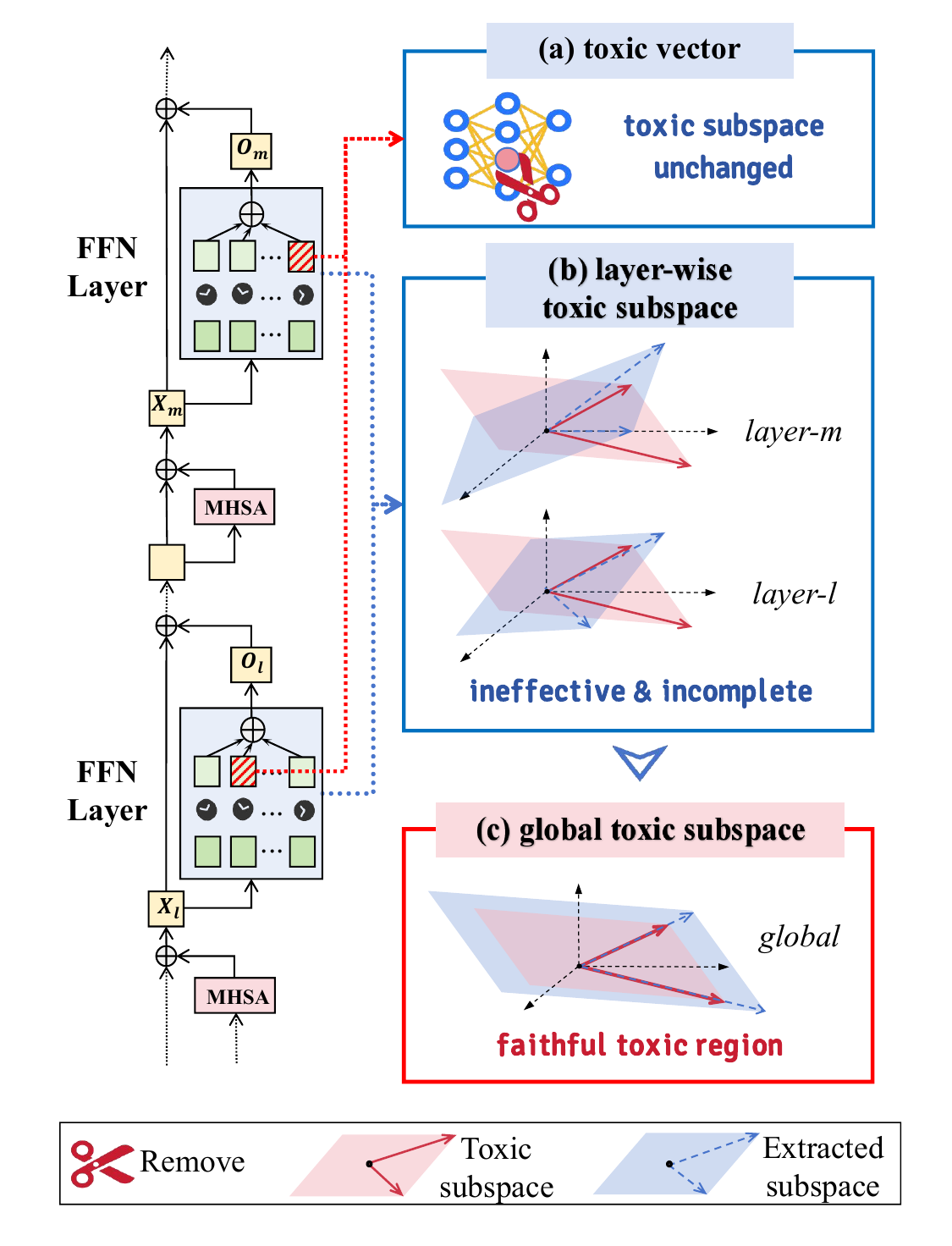}
    \caption{(a) Removing toxic vectors do not alter the underlying toxic subspace. (b) Layer-wise subspaces are limited and fail to capture complete toxic features. (c) Global toxic subspace provides a more faithful  representation of toxic region.}
    \label{fig:introduction}
\end{figure}

However, we find that both views exhibit limitations, as shown in Figure~\ref{fig:introduction}. 
We first observe that suppressing or removing toxic vectors do not effectively reduce toxic outputs. Instead, toxic generation is primarily driven by the cumulative directional bias of FFN outputs toward toxicity ($\mathsection$\ref{sec:Limitations of Toxic Vectors}).
These results motivate us to model the toxic region as a subspace formed by these toxic directions.
% Based on this, we propose that the toxic region is better represented as a subspace formed by toxic directions.
While ProFS emphasizes the value of subspace-level modeling, its layer-wise contrastive extraction fails to identify effective toxic directions at each layer. This is largely due to the varying capacity of FFN to capture toxic features ($\mathsection$\ref{sec:Limitations of Layer-wise Toxic Subspace}), resulting in the extracted subspaces that are often ineffective and incomplete.
Building on these findings, we further observe that toxic directions are shared across layers ($\mathsection$\ref{sec:Global Toxic Subspace}).
Therefore, aggregating them into a unified global toxic subspace provides a more faithful representation of the toxic region.

Motivated by above analysis, we propose a lightweight detoxification method, GloSS (\textbf{\underline{Gl}}obal T\textbf{\underline{o}}xic \textbf{\underline{S}}ubspace \textbf{\underline{S}}uppression), without requiring large-scale data or retraining ($\mathsection$\ref{sec:GloSS}).
GloSS first extracts candidate toxic directions from each layer by applying SVD to activation differences between multiple toxic and non-toxic input pairs.
It then ranks all candidates globally and selects high-scoring ones to ensure that only directions with meaningful toxicity are retained.
Principal components are subsequently extracted from the selected directions to form a unified global toxic subspace.
To suppress toxicity, the value weights of each FFN modules are projected onto the orthogonal complement of this subspace, effectively removing toxic components while preserving the model’s general capabilities.

We evaluate the effectiveness of GloSS through extensive experiments across different LLMs ($\mathsection$\ref{sec:Experiment}). The results demonstrate that:
1) GloSS achieves lower toxicity scores than ProFS and other baselines, while preserving the model’s general capabilities. This supports the conclusion that removing the global toxic subspace enables more effective detoxification.
2) Despite using fewer training samples, both GloSS and ProFS outperform SSFT and DPO, highlighting the effectiveness of safety mechanism based approaches compared to traditional fine-tuning methods.
3) The global toxic subspace exhibits a low-dimensional structure, suggesting that toxicity is concentrated in a compact region of the model’s representation space.

In summary, our contributions are the following:
\begin{itemize}[itemsep=0pt, topsep=2pt] 
  \item  We present a mechanistic understanding of how toxicity emerges in LLMs and identify the global toxic subspace as a more faithful representation of toxic regions.
  \item  We propose GloSS, a lightweight detoxification method that suppresses toxicity via subspace modeling, without requiring additional data or model retraining.
  \item We conduct extensive experiments demonstrating that GloSS achieves state-of-the-art detoxification performance while preserving the general capabilities of LLMs.
\end{itemize}

\begin{figure*}[h]
  \centering
  \subfloat[Enhance activation of vectors]{
    \label{fig:subfig-a}
    \includegraphics[width=0.31\textwidth]{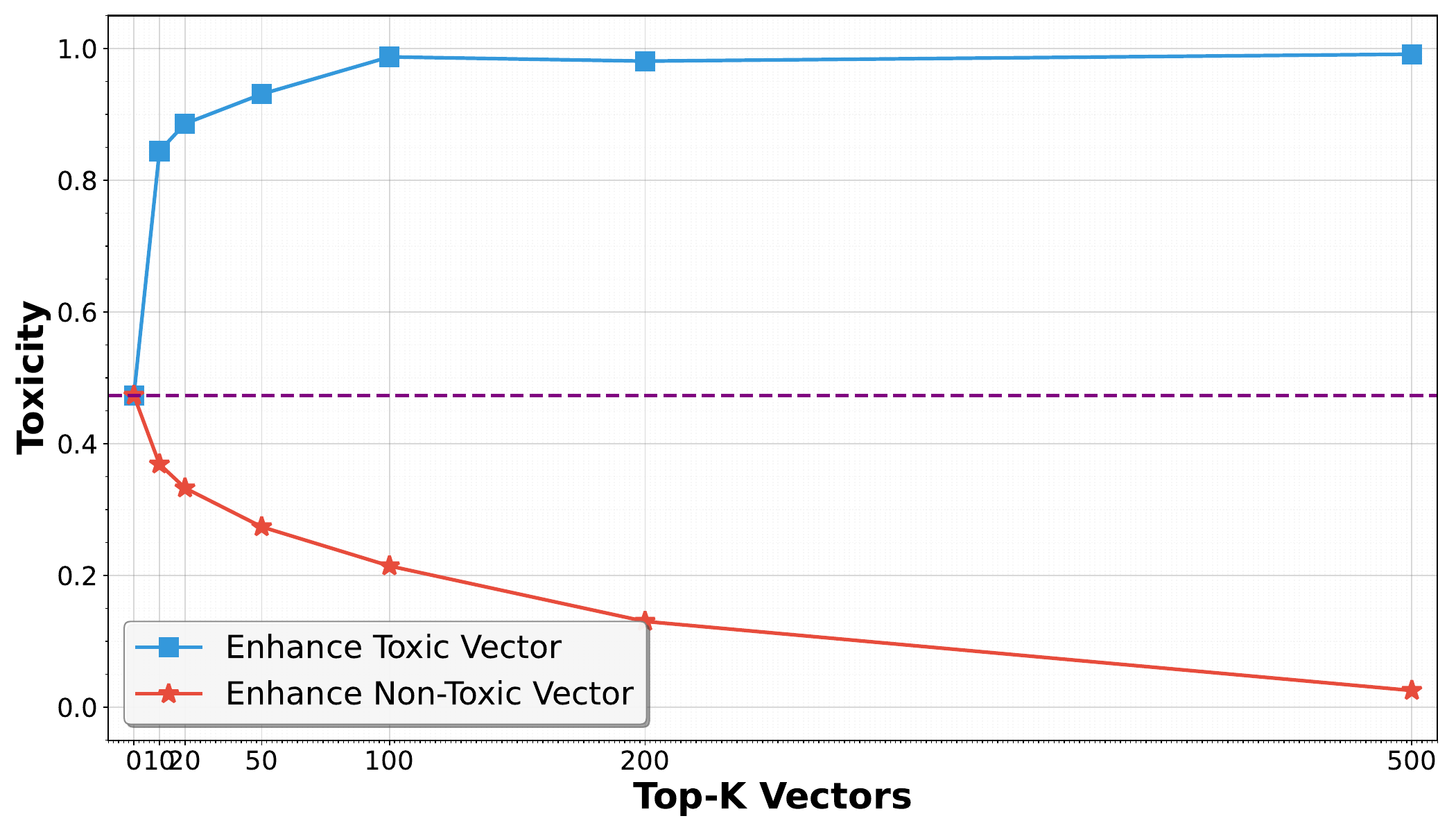}  
    \label{fig:four-panel(a)}
  }
  \hfill
  \subfloat[Suppress activation of vectors]{
    \label{fig:subfig-b}
    \includegraphics[width=0.31\textwidth]{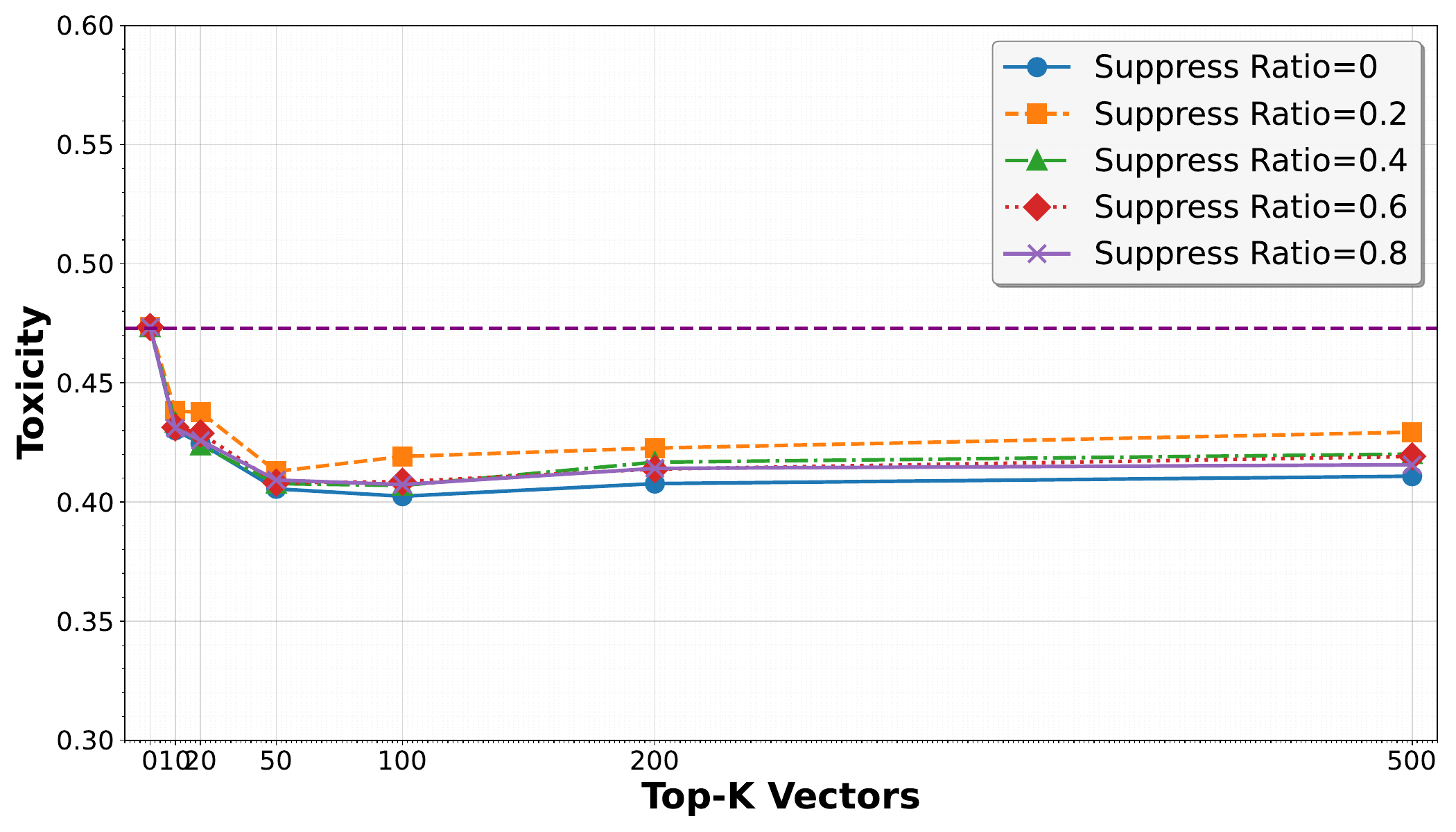}
    \label{fig:four-panel(b)}
  }
  \hfill
  \subfloat[Reverse activation of vectors]{
    \label{fig:subfig-c}
    \includegraphics[width=0.31\textwidth]{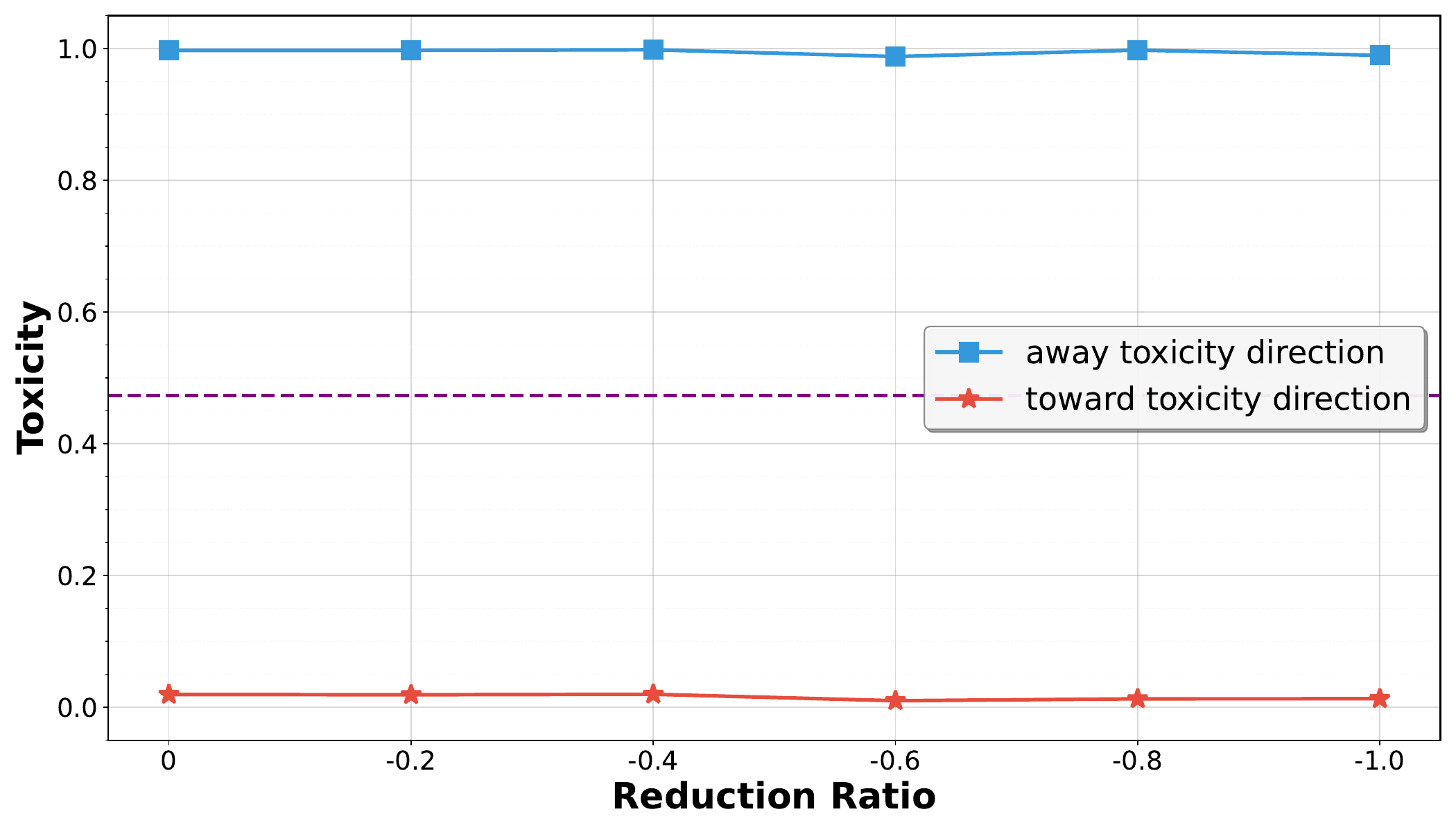}
    \label{fig:four-panel(c)}
  }
  \caption{\textbf{Results of Different Operations on Activation of Vectors.} (a) Enhance different numbers of toxic and non-toxic value vector activations, selectively; (b) Suppress toxic vector activations at different proportions; (c) Reversing value vector activations steers the FFN blocks either toward or away from the toxic direction.}
  \label{fig:four-panel}
\end{figure*}

\section{Preliminaries}

In this section, we introduce the background and define the notations used in our work.

\noindent \textbf{FFN as Linear Combinations of Value Vectors.}
Transformer-based language models are composed of  stacked Transformer layers~\cite{vaswani2017attention}. Each layer includes a Multi-head Self-attention (MHSA) module and a Feed-Forward Network (FFN), both equipped with residual connections and layer normalization.

Given an input sequence $\mathbf{w} = \langle w_0, \dots, w_t \rangle$, the model maps each token $w_i$ to an embedding $\mathbf{e}_i \in \mathbb{R}^d$ using the embedding matrix $E$.
At each layer $\ell$, the FFN receives the hidden state $\mathbf{x}_i^{\ell} \in \mathbb{R}^d$ corresponding to token $i$ and produces an intermediate output $\mathbf{o}_i^{\ell} \in \mathbb{R}^d$. The updated representation after applying the FFN and residual connection is denoted as $\tilde{\mathbf{x}}_i^{\ell} \in \mathbb{R}^d$:
\begin{align}
\mathbf{o}_i^{\ell} &= \mathrm{FFN}^{\ell}(\mathbf{x}_i^{\ell}) \\
\tilde{\mathbf{x}}_i^{\ell} &= \mathbf{x}_i^{\ell} + \mathbf{o}_i^{\ell}
\end{align}

FFN in each Transformer layer typically follows a two-layer structure. It can be interpreted as computing a context-dependent linear combination of learned value vectors~\cite{geva-etal-2022-transformer}. Specifically, the FFN outputs at layer $\ell$ is given by:
\begin{equation}
\begin{aligned}
\mathrm{FFN}^{\ell}(\mathbf{x}^{\ell})
  &= f\left(W_K^{\ell} \mathbf{x}^{\ell}\right) W_V^{\ell} \\
  &= \sum_{i=1}^{d_m} f\left(\mathbf{x}^{\ell} \cdot \mathbf{k}_i^{\ell} \right)\, \mathbf{v}_i^{\ell} = \sum_{i=1}^{d_m} m_i^{\ell}\, \mathbf{v}_i^{\ell}
\end{aligned}
\end{equation}
We focus on the FFN update for a single token and omit the token index for simplicity, i.e., $\mathbf{x}^{\ell} := \mathbf{x}_i^{\ell}$. The weight matrices $W_K^\ell, W_V^\ell \in \mathbb{R}^{d_m \times d}$ parameterize the FFN at layer $\ell$. We denote the $i$-th row of $W_K^\ell$ as $\mathbf{k}_i^\ell$ (key vector) and the $i$-th column of $W_V^\ell$ as $\mathbf{v}_i^\ell$ (value vector). The function $f(\cdot)$ represents a non-linear activation, such as GELU.

FFN outputs can be interpreted as a weighted sum of value vectors $\mathbf{v}_i^\ell$, where each vector is scaled by a context-dependent coefficient $m_i^\ell := f(\mathbf{x}^{\ell} \cdot \mathbf{k}_i^{\ell})$. This shows that FFNs compute a linear combination of learned semantic directions.

\begin{table*}[ht]
\center
\caption{Top Toxic and Non-Toxic Vectors in GPT-2 Projected into Vocabulary Space Under Different Activation. Negative activation of toxic vectors yields non-toxic output, while that of non-toxic vectors can produce toxicity. }
% \textcolor{red}{(Warning: Italicized examples may contain toxic content.)}}
\small
\begin{tabularx}{\textwidth}{@{}c c CC@{}}
\toprule
    \multirow{2}{*}{\textbf{Vector}}
      & \multirow{2}{*}{\textbf{Toxicity}}
      & \multicolumn{2}{c}{\textbf{Top Tokens}} \\
    \cmidrule(lr){3-4}          % 只在第3,4列下画横线
      & 
      & \textbf{Positive activation}
      & \textbf{Negative activation} \\
\midrule
$W_\text{toxic}$ && \textit{c*nt, f*ck, a**hole, d*ck, wh*re, holes} & orate, onding, medium, esp, iations, rece \\

$\text{MLP.v}_{770}^{19}$ & \ding{51} & \textit{sh*t, a**, cr*p, f*ck, c*nt, garbage, trash} & anni, anwhile, Uri, iscovery, GMT, owship \\
$\text{MLP.v}_{771}^{12}$ & \ding{51} & \textit{delusional, hypocritical, arr**nt, no**nse} & toggle, MAP, uration, bis, uala, Mine, Sig \\
$\text{MLP.v}_{2669}^{18}$ & \ding{51} & \textit{degener, whining, idiots, stupid, sm**g} & iment, assetsadobe, ANGE, href, querque\\

$\text{MLP.v}_{1882}^{10}$ & \ding{55} & buoy, stabilized, clud, helps, breaks, shows & \textit{ardo, man**c, bul***it, fu**ing, nonsense}\\
$\text{MLP.v}_{1307}^{11}$ & \ding{55} & aker, atch, encer, erick, wik, follow, participant & \textit{damn, kidding, freaking, darn, p**s, !, booze} \\
$\text{MLP.v}_{3094}^{7}$ & \ding{55} & dialect, texts, staples, rend,  repertoire, sessions & \textit{wasting, ternity, usterity, UCK, closure, fuss} \\
\bottomrule
\end{tabularx}
\label{tab:toxic-vector-top-tokens}
\end{table*}

\noindent \textbf{Interpreting Vectors in Vocabulary Space.} 
To interpret the semantic meaning of a vector $\mathbf{u} \in \mathbb{R}^{d}$ in the embedding space, we project it into the vocabulary space using the output embedding matrix $E = [\mathbf{e}1, \ldots, \mathbf{e}{|\mathcal{V}|}]^\top \in \mathbb{R}^{|\mathcal{V}| \times d}$, where $\mathcal{V}$ denotes the vocabulary~\cite{Geva2020TransformerFL}:
\begin{align}
    r = E \mathbf{u} \in \mathbb{R}^{|\mathcal{V}|}
\label{align:projection}
\end{align}
We select the top-$k$ tokens from the projection of $\mathbf{u}$, offering an interpretable approximation of its semantic content.
\textit{Notably, this projection depends only on the direction of $\mathbf{u}$, not its magnitude.}

\section{Motivation} \label{sec:motivation}

Two main perspectives have been proposed regarding the presence of toxic regions within the FFN module of Transformer:
(1) a set of toxic vectors~\cite{lee2024a}, and (2) layer-wise toxic subspace~\cite{uppaal2025model}. 
Although both frameworks offer valuable insights, our findings suggest that they may not fully capture the underlying mechanisms of toxicity.

To investigate this, we conduct experiments on GPT-2 Medium (henceforth GPT2) using the challenge subset of the REALTOXICITYPROMPTS dataset~\cite{gehman-etal-2020-realtoxicityprompts}, which includes 1,199 prompts designed to elicit highly toxic responses.
Following~\cite{uppaal2025model}, we use Detoxify\footnote{\url{https://github.com/unitaryai/detoxify}} to score the toxicity of the first 10 generated tokens for each prompt.

\subsection{Limitations of Toxic Vectors} \label{sec:Limitations of Toxic Vectors}

~\citet{lee2024a} suggest that toxic region is formed by a set of toxic vectors selected via a trained probe vector. However, this view may be limited.

\begin{tcolorbox}[takeawaysstyle, title=Observation]
Suppressing or removing toxic vectors fails to mitigate toxicity effectively.
\end{tcolorbox}

We begin by examining whether toxic vectors are correlated with toxicity.
To this end, we use a toxic probe vector to identify the most and least similar value vectors in FFN.
We refer to these as toxic and non-toxic vectors, respectively.
During generation, we selectively enhance varying numbers of toxic and non-toxic value vector activations and observe the corresponding changes in output toxicity.
Figure~\ref{fig:four-panel(a)} shows the results when activation are scaled by 10. 
As the number of enhanced activations increases, we observe a clear trend: toxicity increases rapidly  when toxic vectors are amplified and decreases when non-toxic vectors are enhanced.  
These results indicate that toxic vectors contribute to the generation of toxic content, while non-toxic vectors play a suppressive role.

To test whether toxic region is truly composed of toxic vectors, we suppress their activations to simulate their removal and observe model outputs.
As shown in Figure~\ref{fig:four-panel(b)}, suppressing toxic vectors reduces toxicity by less than 0.08, with little overall effect.
Even when more vectors are suppressed, toxicity remains high or even rebounds. 
This observation is similar with findings from~\cite{mayne2024ablationnotenough}. 
In summary, Although enhancing the activation of toxic vectors leads to increased toxicity, suppressing them does not significantly reduce it. This suggests that while toxic vectors are correlated with toxic output, they are unlikely to constitute the structural basis of toxic regions.

\begin{tcolorbox}[takeawaysstyle, title=Assumption]
Toxicity arises when FFN outputs are biased toward toxic directions.
\end{tcolorbox}

To further investigate the structure of toxic regions, we conduct a detailed analysis of vector activations.
We first observe that activations of the toxic vectors significantly influence the expression of toxicity. 
As shown in Table~\ref{tab:toxic-vector-top-tokens}, negative activation of a toxic vector leads to non-toxic output; conversely, negative activation of a non-toxic vector results in toxic output.
This suggests that toxicity depends not only on which vectors are involved but also on how they are activated.

Motivated by above observations, and grounded in the view that FFNs operate as linear combinations of value vectors~\cite{geva-etal-2022-transformer}, we hypothesize that toxicity arises when the FFN outputs is biased toward a specific toxic direction. 
To test this hypothesis, we define the normalized toxic probe vector as the toxic direction and design a contrastive experiment with two settings.
\begin{itemize}
    \item \textit{FFN towards the toxic direction}: Aactivation signs follow the similarity (positive stays positive, negative stays negative);
    \item \textit{FFN away from the toxic direction}: Activation signs are flipped (positive becomes negative, negative becomes positive).
\end{itemize}
As shown in Figure~\ref{fig:four-panel(c)}, when FFN outputs are biased toward the toxic direction, the toxicity score remains high (close to 1.00).
In contrast, when biased away from the toxic direction, the score drops toward 0.
These results support our hypothesis that the cumulative directional bias of FFN layers drives toxic generation. Toxic vectors amplify activations along toxicity-aligned directions, and even after removing some vectors, the remaining ones can still combine to induce toxicity.

\noindent \textbf{Conclusion.}\quad 
% Toxic region is not composed of a set of toxic vectors, but emerge from a subspace formed by toxic directions.
% Effective suppression of toxic generation requires totally removing this subspace to prevent model outputs from aligning with toxic directions.
Toxic vectors correlate with toxicity and increase it when amplified, but suppressing them has little effect. This suggests toxicity arises from a cumulative directional bias in FFN outputs toward a toxic subspace, rather than from individual vectors alone.

\begin{table}[t]
  \centering
  \small
  \caption{Toxicity Projection Results Across Layers. The heuristic scaling factor $\alpha=100$. }
  \begin{tabularx}{0.48\textwidth}{@{}cX@{}}
    \toprule
    \textbf{Vector} & \textbf{Top Tokens} \\
    
    \midrule
    $\mathbf{d}_1$ & ften, Painter, proper, nce, AMY, favour, squared \\
    $\mathbf{d}_2$ & proper, Painter, court, Extrem, Court, squared  \\
    $\mathbf{d}_4$ & \textit{po*p, h**ny, nip**es, kittens, tits, sh*t, s**en} \\
    $\mathbf{d}_{12}$  & \textit{sh*t, f*ck, u**er, bag, weed, yeah, dragon, stab} \\
    $\mathbf{d}_{14}$ & \textit{sh*t, f*ck, F*ck, f*cking, b**ch, d*ck, F*CK} \\
    $\mathbf{d}_{24}$ & B, b, C, S, P, L, p, M, F, T, d, A, R, H, V, D, u \\
    $\mathbf{d}_{24}$ & -, (, and, the, a, ", The, s, in, A, The, S, B, b, C \\

    \midrule
    $\mathbf{x}_1$ & Citiz, mum, Levy, Petr, discrep, Guinea, Sponsor \\
    $\mathbf{x}'_1$ & \textit{sh*t, F*ck, f*ck, st*b, ucker, cision, bi*ch, ser} \\
    $\mathbf{x}_{24}$ & the, and, -, (, a, in, I, to, of, The, A, or, for, that \\
    $\mathbf{x}'_{24}$ & \textit{sh*t, f*ck, ucker, F*ck, god, ard, uck, ass, p*op} \\
    
    \bottomrule
    
  \end{tabularx}
  \label{tab:layer_tokens}
\end{table}

\subsection{Limitations of Layer-wise Toxic Subspace}
\label{sec:Limitations of Layer-wise Toxic Subspace}

Prior works have highlighted the importance of toxic subspace, but offered limited insight.  
ProFS~\cite{uppaal2025model} suggests that the toxic subspace is layer-wise, identifying toxic directions based on differences in FFN outputs between toxic and non-toxic prompts at each layer, and combining these directions to form the subspace.

\begin{tcolorbox}[takeawaysstyle, title=Observation]
Layer-wise extraction fails to effectively identify the toxic subspace in most layers.
\end{tcolorbox}

ProFS proposes that an embedding vector at any Transformer layer can be approximated as a combination of stopwords, toxic component, context component, and noise. To analyze this structure, it applies factor analysis to toxic and non-toxic input pairs at a given layer, modeling the embeddings as:
\begin{align}
x_i^+ &= \underbrace{a_i^+ \mu}_{\text{stopwords}} 
       + \underbrace{B f_i}_{\text{toxic}} 
       + \underbrace{\tilde{B} \tilde{f}_i}_{\text{context}} 
       + \underbrace{u_i^+}_{\text{noise}}, \notag \\
x_i^- &= \underbrace{a_i^- \mu}_{\text{stopwords}} \quad\quad\quad
       + \underbrace{\tilde{B} \tilde{f}_i}_{\text{context}} 
       + \underbrace{u_i^- }_{\text{noise}}
\end{align}

Building on this formulation, we input multiple toxic and non-toxic pairs and construct contrastive matrices at each layer.
We then apply SVD to extract the top one-dimensional direction \(\mathbf{d}_\ell\), which is assumed to represent the toxic direction, and project it into the vocabulary space to examine the top-$k$ tokens.
% 实验结果描述
As shown in Table~\ref{tab:layer_tokens}, projections from middle layers show mostly toxic tokens, whereas those from lower and upper layers do not. 
This suggests that layer-wise toxic directions lack effectiveness and consistency, making the resulting subspaces unreliable.

\begin{tcolorbox}[takeawaysstyle, title=Assumption]
The capacity of FFN blocks to capture toxic features varies across layers.
\end{tcolorbox}

If input pairs differ clearly in toxicity, what causes the failure in layer-wise toxic direction extraction?
We hypothesize that this stems from the variation in how FFN blocks model toxic features.
As shown in Table~\ref{tab:layer_tokens}, the projection results exhibit a clear layer-wise pattern.
In the early layers (e.g., the first and second), the contrast between toxic and non-toxic projections mainly involves context words.
In the final layers, the differences shift toward symbols and stopwords.
Only the middle layers consistently reveal toxic tokens; however, both the intensity and semantics of these tokens vary across layers.
These results suggest that the lower and upper layers encode toxicity differently from the middle layers. Even among the middle layers, toxic features are expressed inconsistently, both in strength and type.
This aligns with prior finding~\cite{Sun_Pickett_Nain_Jones_2025}, potentially reflecting functional differences in FFNs across layers.

\noindent \textbf{Conclusion.}\quad
% Based on the assumption, we argue that contrastive extraction fails to identify effective toxic directions, due to the varying capacity of FFN blocks to model toxicity. 
% This results in a toxic subspace that is both unreliable and incomplete.
Due to the varying capacity of FFN blocks to model toxicity, we found that contractive extraction fails to identify effective toxic directions at each layer. Therefore, toxic subspace is unreliable and inconsistent.

\begin{figure}
    \centering
    \includegraphics[width=0.85\linewidth]{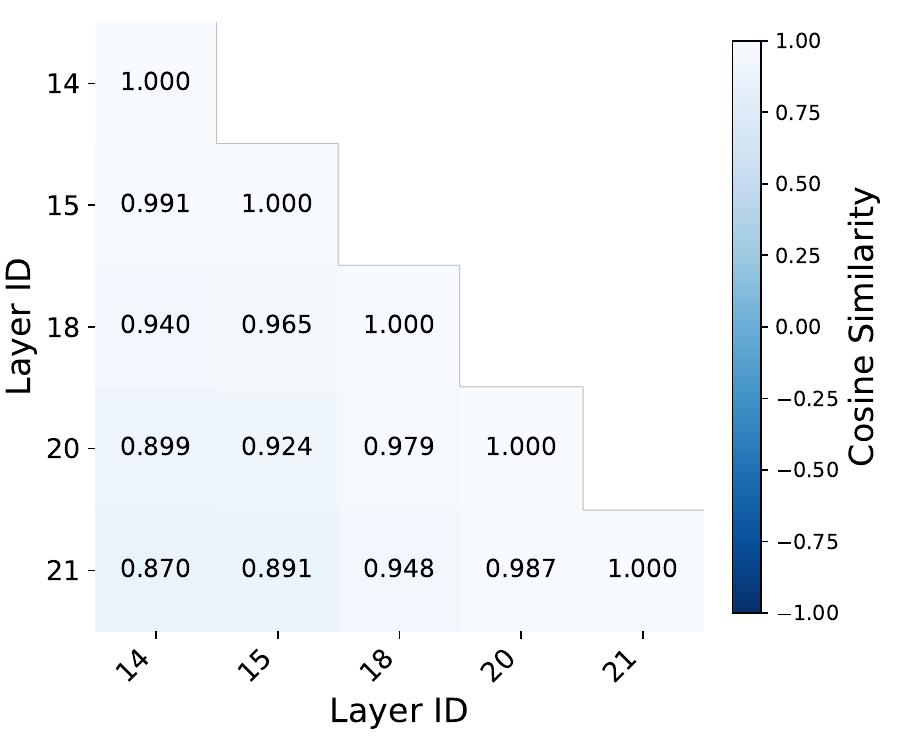}
    \caption{Top-5 Toxic Directions Across Layers. They are primarily located in the middle-to-late layers and exhibit pairwise cosine similarities close to 1.}
    \label{fig:similar}
\end{figure}

\subsection{Global Toxic Subspace}      \label{sec:Global Toxic Subspace}

The toxic region can be viewed as a toxic subspace, but existing layer-wise extraction methods are limited. This raises a key question: how can we model it more effectively?

\begin{tcolorbox}[takeawaysstyle, title=Observation]
The toxic subspace is shared across all layers.
\end{tcolorbox}

We further analyze the directions extracted from each layer in Section~\ref{sec:Limitations of Layer-wise Toxic Subspace} by ranking all candidate directions from different layers using a predefined bad words list $\mathcal{B}$~\cite{gehman-etal-2020-realtoxicityprompts}. Each direction $\mathbf{d}_\ell$ is projected into the vocabulary space, and its top-$m$ tokens $\mathcal{T}_{\mathbf{d}_\ell}$ are compared against $\mathcal{B}$. The toxicity score is computed as: 
\begin{align}
\text{tox\_score}(\mathbf{d}_\ell) = \frac{|\mathcal{T}_{\mathbf{d}_\ell} \cap \mathcal{B}|}{m}
\end{align}
We select the top-5 directions with the highest toxicity scores based on this metric. These directions are mainly concentrated in the middle-to-late layers (e.g., layers 14, 15, 18, 20, and 21) and exhibit high pairwise cosine similarity, as illustrated in Figure~\ref{fig:similar}.

Additionally, we use 1,000 non-toxic WikiText-2~\cite{merity2016pointersentinelmixturemodels} sentences as prompts to compute the average token activation at each layer, denoted as $\mathbf{x}_\ell$.
We then select the top-ranked toxic direction $\mathbf{d}_{\ell_0}$ at layer $\ell_0 = 14$, and shift the average activation along this direction:
\begin{align}
\mathbf{x}_\ell' = \mathbf{x}_\ell + \alpha \cdot \mathbf{d}_{\ell_0}
\end{align}
$\alpha$ is a heuristic scaling factor. As shown in Table~\ref{tab:layer_tokens}, shifting activations along a toxic direction in layers 1 and 24 converts the projected tokens from non-toxic to toxic.
This suggests that toxicity directions are shared across the model, and the subspace they form is therefore global in nature.

\begin{figure*}
    \centering
    \includegraphics[width=\linewidth]{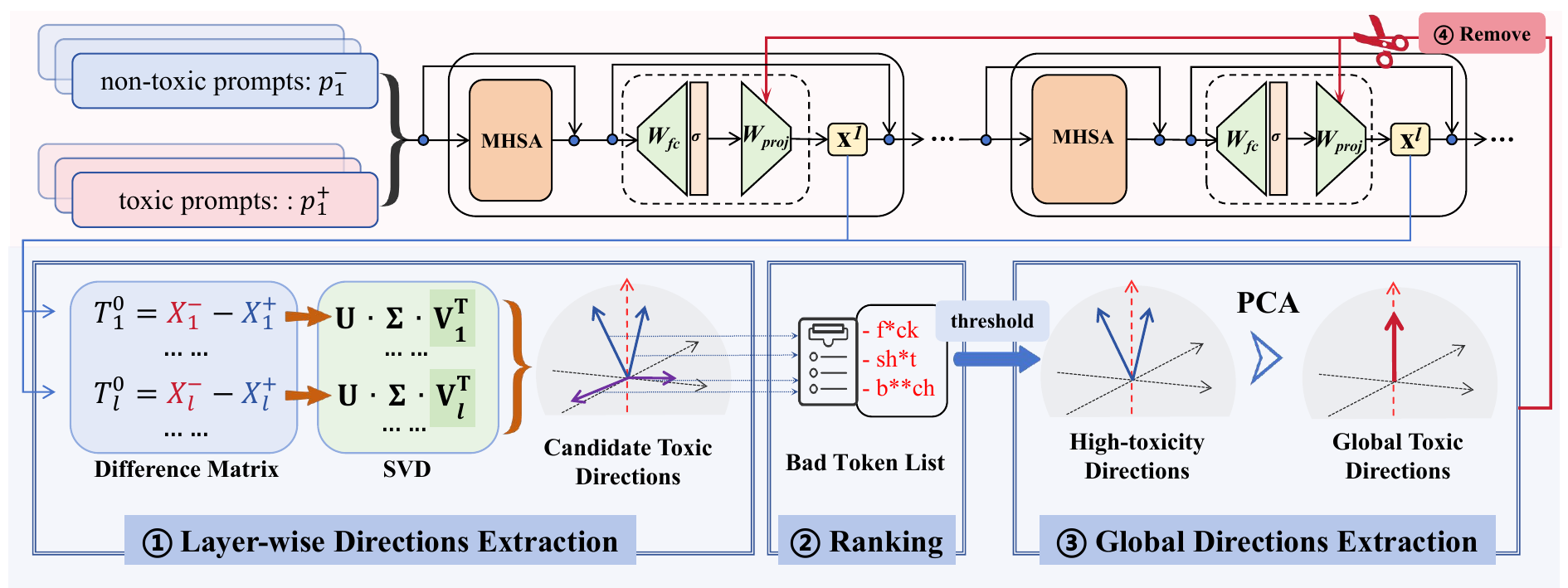}
    \caption{The overview of GloSS. It identifies and removes the global toxic subspace through a four-stage procedure to effectively reduce toxic generation. The intervention is applied by modifying $W_{proj}$ in the FFN modules. }
    \label{fig:gloss}
\end{figure*}

\noindent \textbf{Conclusion.}\quad
The above observations reveal that toxic directions are not limited to individual layers but are consistently shared across multiple layers. We therefore consider the global toxic subspace, constructed by aggregating toxic directions from all layers, to be a more essential representation of toxic regions in the model.

\section{Detoxification Method: GloSS}      \label{sec:GloSS}

% 第一段：承上启下段
% 上一章节介绍了global toxic subspace是一个更本质的毒性区域。我们基于此提出了一种去毒方法。提出一种全局毒性子空间三阶段的提取策略与移除策略，识别并移除模型中潜在的结构性毒性表达区域，从而实现有效的去毒。
Building on the insights from Section~\ref{sec:motivation}, we propose a detoxification method, \textbf{GloSS} (\textbf{\underline{Gl}}obal T\textbf{\underline{o}}xic \textbf{\underline{S}}ubspace \textbf{\underline{S}}uppression), a detoxification method that identifies and removes the global toxic subspace through a four-stage procedure to effectively reduce toxic generation, as shown in Figure~\ref{fig:gloss}.

\noindent \textbf{Step 1: Layer-wise Directions Extraction.} Following ProFS, we identify candidate toxic directions by comparing the FFN output of paired toxic and non-toxic inputs at each layer.
Given a model and $N$ sentence pairs $\mathcal{D}_{\text{pref}} = \{(p_i^+, p_i^-)\}_{i=1}^N$, we compute the average FFN output  at each layer for every input pair, and stack them into matrices $X_\ell^+, X_\ell^- \in \mathbb{R}^{N \times d}$. The initial contrastive representation is then defined as $T_\ell^0 := X_\ell^+ - X_\ell^-$. 
To mitigate the influence of frequent token semantics,we perform mean-centering to obtain a refined contrastive matrix $T_\ell$.

Finally, we apply singular value decomposition (SVD) to $T_\ell$ to extract the dominant directions:
\begin{align}
    \mathbf{U} \boldsymbol{\Sigma} \mathbf{V}_\ell^\top = T_\ell, \quad \mathbf{V}_\ell=(\mathbf{v}_\ell^1, \mathbf{v}_\ell^2, \ldots, \mathbf{v}_\ell^N)
\end{align}
We extract the top-$k$ right singular vectors $\mathbf{v}_\ell^1, \mathbf{v}_\ell^2, \ldots, \mathbf{v}_\ell^k \in \mathbb{R}^{d}$ as the candidate toxic directions at layer $\ell$. 
Larger $k$ values enable capture a richer set of toxic representations.

\noindent \textbf{Step 2: Ranking.} In this step, we rank all candidate toxic directions ${\mathbf{v}}$ extracted from each layers. Each direction is projected into the vocabulary space using the output embedding matrix $E \in \mathbb{R}^{|\mathcal{V}| \times d}$ as described in Equation (\ref{align:projection}).
We then select the top-$m$ tokens from the projection result, denoted as $\mathcal{T}{v}$, and compute the toxicity score by measuring the overlap with a predefined bad words list $\mathcal{B}$, as defined in Section~\ref{sec:Global Toxic Subspace}:
\begin{align}
    \text{tox\_score}(\mathbf{v}) = \frac{|\mathcal{T}_v \cap \mathcal{B}|}{m}
\end{align}
This score quantifies how strongly direction $\mathbf{v}$ is associated with toxicity and serves as the basis for cross-layer ranking.

\begin{table*}
  \centering
  \caption{Comparison of Detoxification Effectiveness and General Capability Across Methods and Models. ProFS and GloSS are trained on $N=500$ pairwise toxic samples, while SSFT and DPO use $N=2000$. Here, $N$ denotes the number of prompt pairs. \textit{Noop} refers to the original model without any modification.}
  \label{tab:main-results}
  \resizebox{\textwidth}{!}{
  \begin{tabular}{c*{4}{cc}}
    \toprule
   \multirow{2}{*}{Methods}
      & \multicolumn{2}{c}{GPT-2 Medium} 
      & \multicolumn{2}{c}{GPT-J 6B}
      & \multicolumn{2}{c}{OPT 6.7B} 
      & \multicolumn{2}{c}{Mistral 7B}  \\
    \cmidrule(lr){2-3} \cmidrule(lr){4-5} \cmidrule(lr){6-7} \cmidrule(lr){8-9} 
      & Toxicity & Perplexity  
      & Toxicity & Perplexity  
      & Toxicity & Perplexity  
      & Toxicity & Perplexity  \\
    \midrule
    Noop 
      & 0.480 & 29.70
      & 0.453 & 13.24    
      & 0.465 & 14.67 
      & 0.425 & 7.49    \\

    SSFT~\cite{ouyang2022training}
      & 0.398 & 30.50
      & 0.429 & 13.18  
      & 0.434 & 14.04
      & 0.417 & 7.34 \\

    DPO~\cite{dpo2023}
      & 0.363 & 29.86 
      & 0.437 & 13.96 
      & 0.453 & 14.37 
      & 0.364 & 7.52  \\

    ProFS~\cite{uppaal2025model}
      & 0.268 & 32.50 
      & 0.374 & 14.53  
      & 0.435 & 13.83 
      & 0.304 & 7.99\\

    % DINM~\cite{wang-etal-2024-detoxifying}
    %   &   --  &   --  
    %   &  --   &  --
    %   &  --   &  --
    %   &  --   &  --  \\

    % ToxRev~\cite{leong-etal-2023-self}
    %   &   --  &   --  
    %   &  --   &  --
    %   &  --   &  --
    %   &  --   &  --  \\

    \textbf{GloSS(ours)}
      & \underline{\textbf{0.208}} & 32.31  
      & \underline{\textbf{0.283}} & 14.52 
      & \underline{\textbf{0.352}} & 14.53 
      & \underline{\textbf{0.271}} & 7.95 \\
    \bottomrule
  \end{tabular}
  }
\end{table*}

\noindent \textbf{Step 3: Global Toxic Directions Extraction.} To identify high-confidence toxic directions across all layers, we define a threshold $\tau$ based on the distribution of toxicity scores $\text{tox\_score}(\mathbf{v})$:
\begin{align}
\tau = \mu + \alpha \cdot \sigma
\end{align}
Here, $\mu$ and $\sigma$ are the mean and standard deviation of the toxicity scores, respectively. $\alpha$ is a scaling parameter that controls the selection strictness. 
Accordingly, we select directions whose toxicity scores exceed this threshold.
\begin{align}
\mathcal{V}_{\text{high}} = \{\mathbf{v}_i \mid \text{tox\_score}(\mathbf{v}_i) > \tau\}
\end{align}
This subset $\mathcal{V}_{\text{high}}$ captures the most salient directions associated with toxic content across layers.

To extract the principal components from $\mathcal{V}_{\text{high}}$, we apply PCA and retain the minimal number of components whose cumulative explained variance exceeds a threshold $\eta$:
\begin{align}
\mathbf{V}_{\text{PCA}} = \text{PCA}{\geq \eta}(\mathcal{V}_{\text{high}}) \in \mathbb{R}^{r \times d}
\end{align}
The resulting matrix $\mathbf{V}_{\text{PCA}}$ contains the dominant directions that best represent toxicity signals consistently shared across layers.

\noindent \textbf{Step 4: Removing.}
We mitigate toxic representations by projecting the model’s parameters onto the orthogonal complement of the global toxic subspace.
Given the $n$ orthonormal global toxic directions $\{\mathbf{d}_1, \mathbf{d}_2, \dots, \mathbf{d}_n\}$ from $\mathbf{V}_{\text{PCA}}$,we define the projection matrix for the toxic subspace as:
\begin{align}
\mathbf{P}^{\text{toxic}} = \sum_{i=1}^{n} \mathbf{d}_i \mathbf{d}_i^\top
\end{align}

To suppress toxicity, we apply the projection to the FFN value matrices $W_{V,\ell}$ at each layer $\ell$:
\begin{align}
W_{V,\ell}^{\text{proj}} = \left(\mathbf{I} - \mathbf{P}^{\text{toxic}}\right) W_{V,\ell}^{\text{orig}}
\end{align}
This operation removes toxic components while preserving semantic content, enabling lightweight, interpretable detoxification without retraining or performance loss.

\begin{table*}
  \centering
  \caption{Dimensionality of Toxic Subspace Identified by GloSS. The subspace generally covers less than 1\% of the hidden space, and its most toxic directions primarily correspond to toxic tokens in the vocabulary.}
  \label{tab:low}
  \resizebox{\textwidth}{!}{
  \begin{tabular}{c c c c c c}
    \toprule
    % 三列单独占两行
    \multirow{2}{*}{\textbf{Model}} &
      \multirow{2}{*}{\textbf{tox\_dim}} &
      \multirow{2}{*}{\textbf{n\_hidden}} &
      \multirow{2}{*}{\textbf{Ratio}} &
      \multicolumn{2}{c}{\textbf{Projection}} \\  % Projection 占一行两列
    \cmidrule(lr){5-6}
    % 第二行只填 Projection 的子标题
    & & & & \textbf{Direction} & \textbf{Top Tokens} \\
    \midrule
    \multirow{2}{*}{GPT-2 Medium}   & \multirow{2}{*}{4} & \multirow{2}{*}{1024} & \multirow{2}{*}{0.004} & $\mathbf{d}_1$ & \textit{f**ked, really, sh*t, kinda, da*n, f**king, crazy} \\
    & & & & $\mathbf{d}_2$ & \textit{p*ss, st**id, upid, F*ck, ass**le, p**sed, godd} \\

    \multirow{2}{*}{GPT-J 6B}   & \multirow{2}{*}{5} & \multirow{2}{*}{4096} & \multirow{2}{*}{0.001} & $\mathbf{d}_1$ & f**kin, f*cking, albums, album, f*ck, peaked \\
    & & & & $\mathbf{d}_2$ & males, Se*ual, vag**al, Males, Sex, Females \\

    \multirow{2}{*}{OPT 6.7B}   & \multirow{2}{*}{21} & \multirow{2}{*}{4096} & \multirow{2}{*}{0.005} & $\mathbf{d}_1$ & f*ck, sh*t, p*ss, b**ch, f*cking, f*cked, as**ole \\
    & & & & $\mathbf{d}_2$ & Male, male, r*ped, female, Female, sex**lly \\

    \multirow{2}{*}{Mistral 7B}   & \multirow{2}{*}{45} & \multirow{2}{*}{4096} & \multirow{2}{*}{0.011} & $\mathbf{d}_1$ & se**al, s*x, p*rn, pen*s, r*pe, actor, biological \\
    & & & & $\mathbf{d}_2$ & f*cking, f*ck, c*ck, UK, f*cked, sh*t, d*ck, rack \\

    \bottomrule
  \end{tabular}}
\end{table*}

\section{Experiment}    \label{sec:Experiment}

\subsection{Experiment Setup}

\noindent \textbf{Base LLMs.}\quad
Our experiments on four large language models of varying sizes, including GPT-2 Medium~\cite{radford2019language}, GPT-J(6B)~\cite{wang2021gpt}, OPT-6.7B~\cite{zhang2022opt}, and Mistral-7B~\cite{jiang2024identifying}.

\noindent \textbf{Baseline Methods.}\quad
We compare our method against several baselines, including SSFT~\cite{ouyang2022training}, DPO~\cite{dpo2023}, and ProFS~\cite{uppaal2025model}. The implementation details are shown in ($\mathsection$~\ref{sec:Experimental Detail}).

\noindent \textbf{Evaluation.}\quad
We evaluate both the toxicity and the general capabilities of the model. To assess toxicity, we use the challenge subset of the REALTOXICITYPROMPTS~\cite{gehman-etal-2020-realtoxicityprompts} dataset as input prompts and measure the toxicity of generated responses using Detoxify. 
To evaluate general capabilities, we follow the approach of~\citet{yang2024butterflyeffectmodelediting} and report perplexity on the WikiText-2 validation set~\cite{merity2016pointersentinelmixturemodels}.

\begin{figure}
    \centering
    \includegraphics[width=1\linewidth]{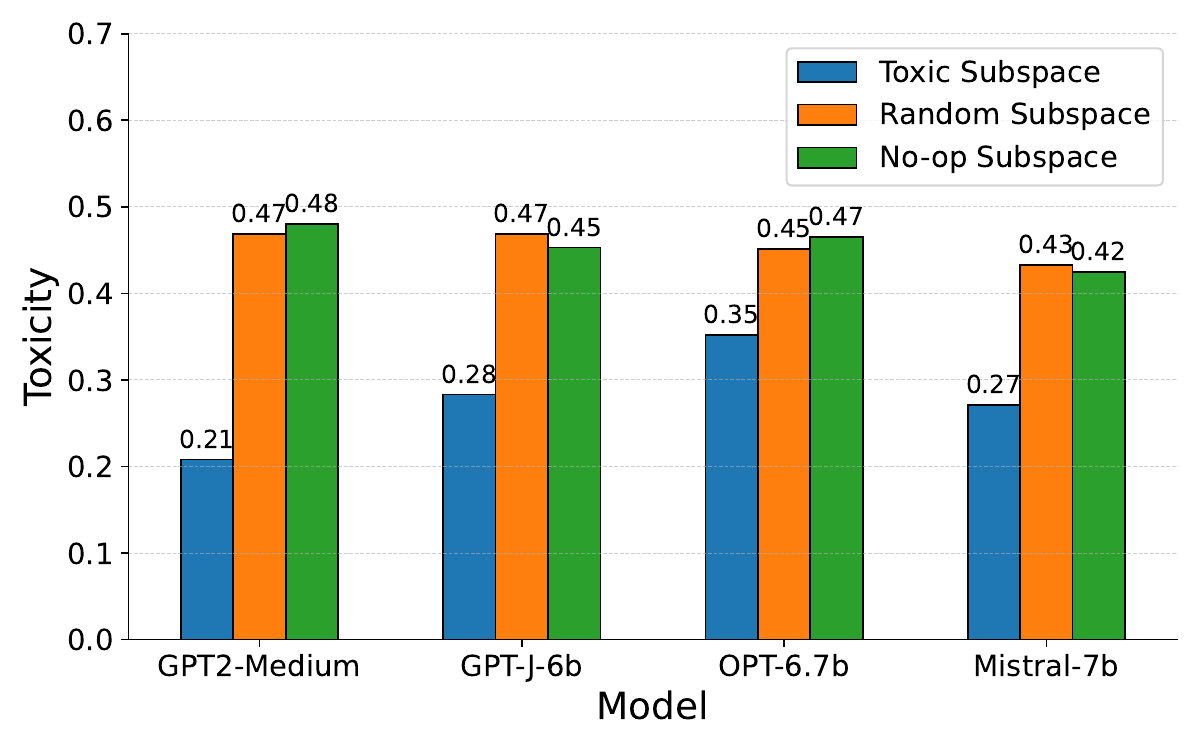}
    \caption{Effectiveness of Extracted vs. Random Subspaces in Toxicity Reduction. \textit{Noop} denotes the original model without any modification.}
    \label{fig:comparison}
\end{figure}

\subsection{Experiment Results}

\noindent \textbf{GloSS Demonstrates Stronger Detoxification with Comparable Model Capability.}
As shown in Table~\ref{tab:main-results},  GloSS maintains stable perplexity scores, indicating that the model’s general language capabilities are not compromised.
In terms of detoxification, GloSS achieves lower toxicity than ProFS, demonstrating the advantage of modeling a global toxic subspace over layer-wise subspaces for capturing and suppressing toxic behaviors.
Moreover, although using only $N=500$ training pairs, which is substantially fewer than the $N=2000$ used by SSFT and DPO, both GloSS and ProFS outperform these fine-tuning based methods in reducing toxicity.
These findings underscore the effectiveness of safety mechanism based approaches over traditional fine-tuning in mitigating toxic outputs.

\noindent \textbf{Global Toxic Subspace is Crucial and Exhibits Low-dimensional Properties.}
We first validate the role of the extracted global toxic subspace in detoxification through a control experiment.
For each model, we construct random subspaces that are orthogonal to the global toxic subspace and have the same dimensionality. These subspaces are then removed from the corresponding FFN layers, and the resulting toxicity levels are compared.
As shown in Figure~\ref{fig:comparison}, removing random subspaces has minimal impact on toxicity reduction and, in some cases, even increases toxicity relative to the original model.
These results confirm that the extracted global toxic subspace captures essential directions specifically associated with toxic behavior.

We further analyze the properties of global toxic subspace and find that it exhibits low-dimensional characteristics.
As shown in Table~\ref{tab:low}, the toxic subspace identified by GloSS spans less than 1\% of the full representation space, and in most cases, remains below 0.5\%.
This suggests that toxic information is concentrated in a small number of directions, supporting the notion of a low-dimensional toxic structure.
Moreover, when the most toxic directions are projected into the vocabulary space, they consistently align with toxic tokens.

\begin{figure}
  \centering
  \subfloat[Toxicity Across Layers in Different LLMs]{
    \label{fig:toxic-a}
    \includegraphics[width=0.9\linewidth]{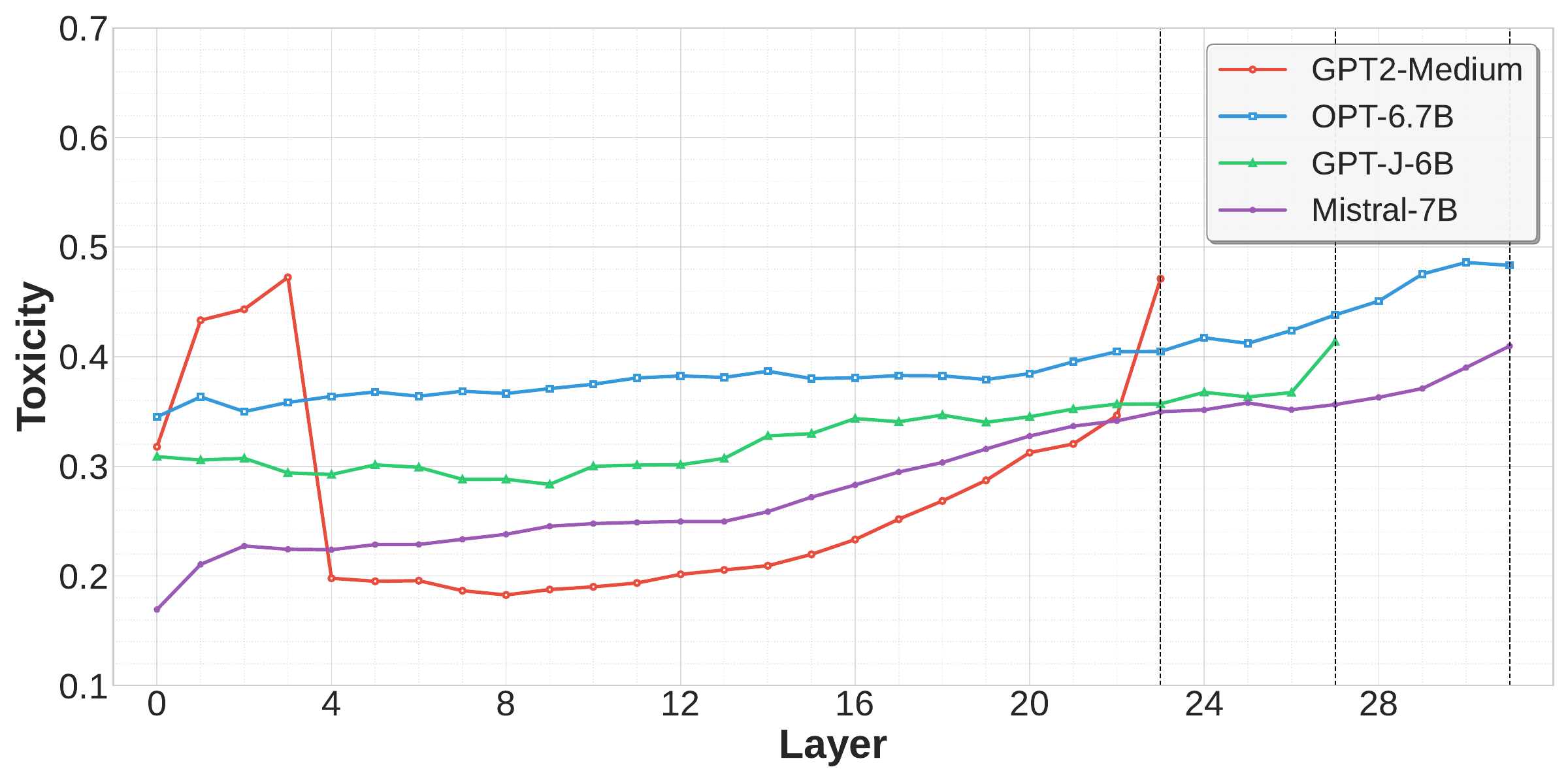}  
  }
  \hfill
  \subfloat[Perplexity Across Layers in Different LLMs]{
    \label{fig:toxic-b}
        \includegraphics[width=0.9\linewidth]{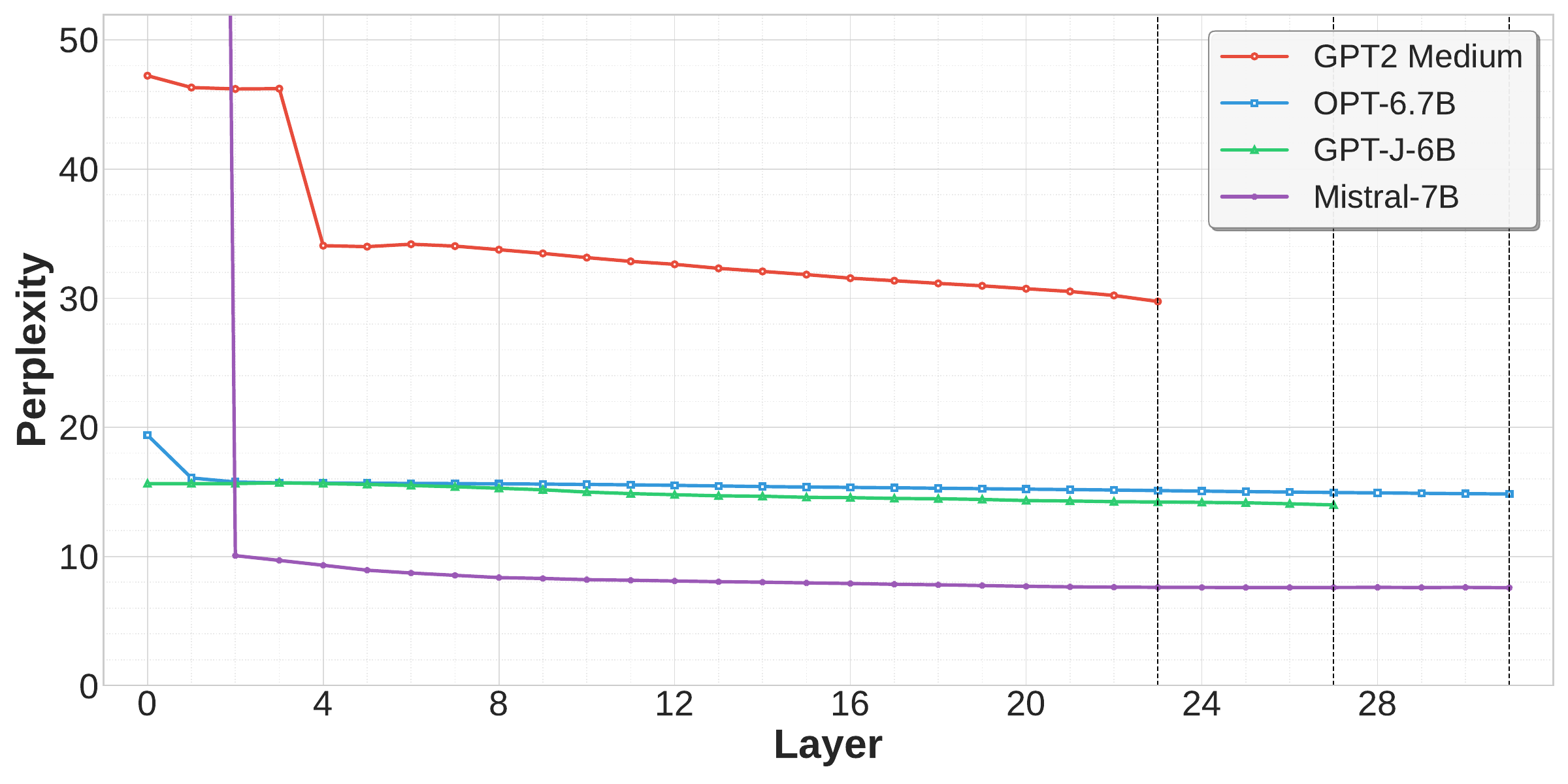}
  }
  \caption{Impact of Projection Layers on Toxicity and Perplexity. (a) Fewer projected layers lead to higher toxicity. (b) Perplexity decreases overall, with a sharp drop when projection is applied to early layers.}
  \label{fig:diff_layer}
\end{figure}

\noindent \textbf{Projection Effects of Different Layers.}
Although the toxic subspace is shared across layers, applying projection at all layers simultaneously can significantly impair model performance.
To investigate this, we systematically evaluate the effects of applying projection starting from different layers up to the final layer, measuring both toxicity and perplexity across four LLMs.
As shown in Figure~\ref{fig:diff_layer}, we find that in all models except GPT-2, reducing the number of projected layers leads to a gradual increase in toxicity and a corresponding decrease in perplexity.
Furthermore, applying projection at early layers causes a sharp drop in perplexity, indicating substantial performance degradation.
For example, in Mistral-7B, projection from layer 2 yields a perplexity of 231.7, while starting from layer 3 reduces it to 9.7, highlighting the model’s sensitivity to early-layer interventions.

\section{Conclusion}

In this work, we propose a mechanistic perspective on toxicity in LLMs and identify the global toxic subspace as a faithful representation of toxic region. Building on this, we introduce GloSS, a lightweight, training-free method that mitigates toxicity by removing toxic subspace from FFN parameters. Our results demonstrate the effectiveness of structural interventions in enhancing LLM safety.

\section{Limitations}

While this paper investigates the underlying mechanisms of toxicity generation in LLMs and proposes an effective detoxification approach, several limitations remain.
First, our evaluation is limited to a small set of open-source LLMs ranging from 0.6B to 7B parameters. The generalization of GloSS to larger models remains to be explored.
Second, we compare GloSS primarily against representative fine-tuning methods (SSFT and DPO). While these baselines are strong and relevant, a broader set of detoxification methods, including prompt-based or detection-based approaches, should also be considered for a more comprehensive evaluation.

\section{Ethics Statement}

This paper focuses on improving the safety of large language models (LLMs) by identifying and suppressing toxic subspaces through interpretable, training-free interventions.
All toxic prompts used for evaluation are sourced from public datasets and manually reviewed to minimize potential harm.
No private or user-generated data is used, and the proposed method does not require model retraining.
We acknowledge potential misuse of internal model insights and take care to present our findings with the goal of strengthening LLM defenses, not enabling harmful applications.

\bibliography{custom}

\begin{thebibliography}{45}
\providecommand{\natexlab}[1]{#1}

\bibitem[{Brown et~al.(2020)Brown, Mann, Ryder, Subbiah, Kaplan, Dhariwal, Neelakantan, Shyam, Sastry, Askell, Agarwal, Herbert-Voss, Krueger, Henighan, Child, Ramesh, Ziegler, Wu, Winter, Hesse, Chen, Sigler, Litwin, Gray, Chess, Clark, Berner, McCandlish, Radford, Sutskever, and Amodei}]{brown2020languagemodelsfewshotlearners}
Tom~B. Brown, Benjamin Mann, Nick Ryder, Melanie Subbiah, Jared Kaplan, Prafulla Dhariwal, Arvind Neelakantan, Pranav Shyam, Girish Sastry, Amanda Askell, Sandhini Agarwal, Ariel Herbert-Voss, Gretchen Krueger, Tom Henighan, Rewon Child, Aditya Ramesh, Daniel~M. Ziegler, Jeffrey Wu, Clemens Winter, and 12 others. 2020.
\newblock \href {https://arxiv.org/abs/2005.14165} {Language models are few-shot learners}.
\newblock \emph{Preprint}, arXiv:2005.14165.

\bibitem[{Dai et~al.(2022)Dai, Dong, Hao, Sui, Chang, and Wei}]{dai-etal-2022-knowledge}
Damai Dai, Li~Dong, Yaru Hao, Zhifang Sui, Baobao Chang, and Furu Wei. 2022.
\newblock \href {https://doi.org/10.18653/v1/2022.acl-long.581} {Knowledge neurons in pretrained transformers}.
\newblock In \emph{Proceedings of the 60th Annual Meeting of the Association for Computational Linguistics (Volume 1: Long Papers)}, pages 8493--8502, Dublin, Ireland. Association for Computational Linguistics.

\bibitem[{Deng et~al.(2024)Deng, Wei, Pang, Ding, Shen, and Cheng}]{deng2024everything}
Jingcheng Deng, Zihao Wei, Liang Pang, Hanxing Ding, Huawei Shen, and Xueqi Cheng. 2024.
\newblock Everything is editable: Extend knowledge editing to unstructured data in large language models.
\newblock \emph{arXiv preprint arXiv:2405.15349}.

\bibitem[{Duan et~al.(2025)Duan, Duan, Yin, Shen, Jing, Zhang, Shen, and Cheng}]{duan2025related}
Zenghao Duan, Wenbin Duan, Zhiyi Yin, Yinghan Shen, Shaoling Jing, Jie Zhang, Huawei Shen, and Xueqi Cheng. 2025.
\newblock Related knowledge perturbation matters: Rethinking multiple pieces of knowledge editing in same-subject.
\newblock \emph{arXiv preprint arXiv:2502.06868}.

\bibitem[{Elhage et~al.(2021)Elhage, Nanda, Olsson, Henighan, Joseph, Mann, Askell, Bai, Chen, Conerly et~al.}]{elhage2021mathematical}
Nelson Elhage, Neel Nanda, Catherine Olsson, Tom Henighan, Nicholas Joseph, Ben Mann, Amanda Askell, Yuntao Bai, Anna Chen, Tom Conerly, and 1 others. 2021.
\newblock A mathematical framework for transformer circuits.
\newblock \emph{Transformer Circuits Thread}, 1(1):12.

\bibitem[{Gehman et~al.(2020)Gehman, Gururangan, Sap, Choi, and Smith}]{gehman-etal-2020-realtoxicityprompts}
Samuel Gehman, Suchin Gururangan, Maarten Sap, Yejin Choi, and Noah~A. Smith. 2020.
\newblock \href {https://doi.org/10.18653/v1/2020.findings-emnlp.301} {{R}eal{T}oxicity{P}rompts: Evaluating neural toxic degeneration in language models}.
\newblock In \emph{Findings of the Association for Computational Linguistics: EMNLP 2020}, pages 3356--3369, Online. Association for Computational Linguistics.

\bibitem[{Geva et~al.(2022)Geva, Caciularu, Wang, and Goldberg}]{geva-etal-2022-transformer}
Mor Geva, Avi Caciularu, Kevin Wang, and Yoav Goldberg. 2022.
\newblock \href {https://doi.org/10.18653/v1/2022.emnlp-main.3} {Transformer feed-forward layers build predictions by promoting concepts in the vocabulary space}.
\newblock In \emph{Proceedings of the 2022 Conference on Empirical Methods in Natural Language Processing}, pages 30--45, Abu Dhabi, United Arab Emirates. Association for Computational Linguistics.

\bibitem[{Geva et~al.(2020)Geva, Schuster, Berant, and Levy}]{Geva2020TransformerFL}
Mor Geva, R.~Schuster, Jonathan Berant, and Omer Levy. 2020.
\newblock Transformer feed-forward layers are key-value memories.
\newblock \emph{ArXiv}, abs/2012.14913.

\bibitem[{Gu et~al.(2025)Gu, Jiang, Shi, Tan, Zhai, Xu, Li, Shen, Ma, Liu, Wang, Zhang, Wang, Gao, Ni, and Guo}]{gu2025surveyllmasajudge}
Jiawei Gu, Xuhui Jiang, Zhichao Shi, Hexiang Tan, Xuehao Zhai, Chengjin Xu, Wei Li, Yinghan Shen, Shengjie Ma, Honghao Liu, Saizhuo Wang, Kun Zhang, Yuanzhuo Wang, Wen Gao, Lionel Ni, and Jian Guo. 2025.
\newblock \href {https://arxiv.org/abs/2411.15594} {A survey on llm-as-a-judge}.
\newblock \emph{Preprint}, arXiv:2411.15594.

\bibitem[{Hallinan et~al.(2022)Hallinan, Liu, Choi, and Sap}]{hallinan2022detoxifying}
Skyler Hallinan, Alisa Liu, Yejin Choi, and Maarten Sap. 2022.
\newblock Detoxifying text with marco: Controllable revision with experts and anti-experts.
\newblock \emph{arXiv preprint arXiv:2212.10543}.

\bibitem[{Hu et~al.(2021)Hu, Shen, Wallis, Allen{-}Zhu, Li, Wang, and Chen}]{DBLP:journals/corr/abs-2106-09685}
Edward~J. Hu, Yelong Shen, Phillip Wallis, Zeyuan Allen{-}Zhu, Yuanzhi Li, Shean Wang, and Weizhu Chen. 2021.
\newblock \href {https://arxiv.org/abs/2106.09685} {Lora: Low-rank adaptation of large language models}.
\newblock \emph{CoRR}, abs/2106.09685.

\bibitem[{Jiang(2024)}]{jiang2024identifying}
Fengqing Jiang. 2024.
\newblock Identifying and mitigating vulnerabilities in llm-integrated applications.
\newblock Master's thesis, University of Washington.

\bibitem[{Keskar et~al.(2019)Keskar, McCann, Varshney, Xiong, and Socher}]{keskar2019ctrl}
Nitish~Shirish Keskar, Bryan McCann, Lav~R Varshney, Caiming Xiong, and Richard Socher. 2019.
\newblock Ctrl: A conditional transformer language model for controllable generation.
\newblock \emph{arXiv preprint arXiv:1909.05858}.

\bibitem[{Korbak et~al.(2023)Korbak, Shi, Chen, Bhalerao, Buckley, Phang, Bowman, and Perez}]{korbak2023pretraining}
Tomasz Korbak, Kejian Shi, Angelica Chen, Rasika~Vinayak Bhalerao, Christopher Buckley, Jason Phang, Samuel~R Bowman, and Ethan Perez. 2023.
\newblock Pretraining language models with human preferences.
\newblock In \emph{International Conference on Machine Learning}, pages 17506--17533. PMLR.

\bibitem[{Lee et~al.(2024)Lee, Bai, Pres, Wattenberg, Kummerfeld, and Mihalcea}]{lee2024a}
Andrew Lee, Xiaoyan Bai, Itamar Pres, Martin Wattenberg, Jonathan~K. Kummerfeld, and Rada Mihalcea. 2024.
\newblock \href {https://openreview.net/forum?id=dBqHGZPGZI} {A mechanistic understanding of alignment algorithms: A case study on {DPO} and toxicity}.
\newblock In \emph{Forty-first International Conference on Machine Learning}.

\bibitem[{Leong et~al.(2023)Leong, Cheng, Wang, Wang, and Li}]{leong-etal-2023-self}
Chak~Tou Leong, Yi~Cheng, Jiashuo Wang, Jian Wang, and Wenjie Li. 2023.
\newblock \href {https://doi.org/10.18653/v1/2023.emnlp-main.269} {Self-detoxifying language models via toxification reversal}.
\newblock In \emph{Proceedings of the 2023 Conference on Empirical Methods in Natural Language Processing}, pages 4433--4449, Singapore. Association for Computational Linguistics.

\bibitem[{Ma et~al.(2025)Ma, Gao, Wang, Wang, Wang, Sun, Ding, Xu, Chen, Zhao, Huang, Li, Zhang, Zheng, Bai, Wu, Qiu, Zhang, Li, Han, Li, Sun, Wang, Gu, Wu, Chen, Zhang, Liu, Gong, Liu, Pan, Xie, Pang, Dong, Jia, Zhang, Ma, Zhang, Gong, Xiao, Erfani, Baldwin, Li, Sugiyama, Tao, Bailey, and Jiang}]{ma2025safetyscalecomprehensivesurvey}
Xingjun Ma, Yifeng Gao, Yixu Wang, Ruofan Wang, Xin Wang, Ye~Sun, Yifan Ding, Hengyuan Xu, Yunhao Chen, Yunhan Zhao, Hanxun Huang, Yige Li, Jiaming Zhang, Xiang Zheng, Yang Bai, Zuxuan Wu, Xipeng Qiu, Jingfeng Zhang, Yiming Li, and 28 others. 2025.
\newblock \href {https://arxiv.org/abs/2502.05206} {Safety at scale: A comprehensive survey of large model safety}.
\newblock \emph{Preprint}, arXiv:2502.05206.

\bibitem[{Mayne et~al.(2024)Mayne, Yang, Mahdi, and Sondej}]{mayne2024ablationnotenough}
Harry Mayne, Yushi Yang, Adam Mahdi, and Filip Sondej. 2024.
\newblock \href {https://arxiv.org/abs/2411.06424} {Ablation is not enough to emulate dpo: How neuron dynamics drive toxicity reduction}.
\newblock \emph{Preprint}, arXiv:2411.06424.

\bibitem[{Meng et~al.(2022)Meng, Bau, Andonian, and Belinkov}]{meng2022locating}
Kevin Meng, David Bau, Alex Andonian, and Yonatan Belinkov. 2022.
\newblock Locating and editing factual associations in gpt.
\newblock \emph{Advances in neural information processing systems}, 35:17359--17372.

\bibitem[{Merity et~al.(2016)Merity, Xiong, Bradbury, and Socher}]{merity2016pointersentinelmixturemodels}
Stephen Merity, Caiming Xiong, James Bradbury, and Richard Socher. 2016.
\newblock \href {https://arxiv.org/abs/1609.07843} {Pointer sentinel mixture models}.
\newblock \emph{Preprint}, arXiv:1609.07843.

\bibitem[{Ou et~al.(2025)Ou, Yao, Zhang, Jin, Sun, Deng, Li, and Chen}]{ou2025llms}
Yixin Ou, Yunzhi Yao, Ningyu Zhang, Hui Jin, Jiacheng Sun, Shumin Deng, Zhenguo Li, and Huajun Chen. 2025.
\newblock How do llms acquire new knowledge? a knowledge circuits perspective on continual pre-training.
\newblock \emph{arXiv preprint arXiv:2502.11196}.

\bibitem[{Ouyang et~al.(2022)Ouyang, Wu, Jiang, Almeida, Wainwright, Mishkin, Zhang, Agarwal, Slama, Ray et~al.}]{ouyang2022training}
Long Ouyang, Jeffrey Wu, Xu~Jiang, Diogo Almeida, Carroll Wainwright, Pamela Mishkin, Chong Zhang, Sandhini Agarwal, Katarina Slama, Alex Ray, and 1 others. 2022.
\newblock Training language models to follow instructions with human feedback.
\newblock \emph{Advances in neural information processing systems}, 35:27730--27744.

\bibitem[{Pan et~al.(2025)Pan, Liu, Chen, Zhou, Yu, and Jia}]{pan2025hidden}
Wenbo Pan, Zhichao Liu, Qiguang Chen, Xiangyang Zhou, Haining Yu, and Xiaohua Jia. 2025.
\newblock The hidden dimensions of llm alignment: A multi-dimensional safety analysis.
\newblock \emph{arXiv preprint arXiv:2502.09674}.

\bibitem[{Qin et~al.(2020)Qin, Shwartz, West, Bhagavatula, Hwang, Bras, Bosselut, and Choi}]{qin2020back}
Lianhui Qin, Vered Shwartz, Peter West, Chandra Bhagavatula, Jena Hwang, Ronan~Le Bras, Antoine Bosselut, and Yejin Choi. 2020.
\newblock Back to the future: Unsupervised backprop-based decoding for counterfactual and abductive commonsense reasoning.
\newblock \emph{arXiv preprint arXiv:2010.05906}.

\bibitem[{Radford et~al.(2019)Radford, Wu, Child, Luan, Amodei, Sutskever et~al.}]{radford2019language}
Alec Radford, Jeffrey Wu, Rewon Child, David Luan, Dario Amodei, Ilya Sutskever, and 1 others. 2019.
\newblock Language models are unsupervised multitask learners.
\newblock \emph{OpenAI blog}, 1(8):9.

\bibitem[{Rafailov et~al.(2023)Rafailov, Sharma, Mitchell, Ermon, Manning, and Finn}]{dpo2023}
Rafael Rafailov, Archit Sharma, Eric Mitchell, Stefano Ermon, Christopher~D. Manning, and Chelsea Finn. 2023.
\newblock Direct preference optimization: your language model is secretly a reward model.
\newblock In \emph{Proceedings of the 37th International Conference on Neural Information Processing Systems}, NIPS '23, Red Hook, NY, USA. Curran Associates Inc.

\bibitem[{Suau et~al.(2024)Suau, Delobelle, Metcalf, Joulin, Apostoloff, Zappella, and Rodríguez}]{suau2024whisperingexpertsneuralinterventions}
Xavier Suau, Pieter Delobelle, Katherine Metcalf, Armand Joulin, Nicholas Apostoloff, Luca Zappella, and Pau Rodríguez. 2024.
\newblock \href {https://arxiv.org/abs/2407.12824} {Whispering experts: Neural interventions for toxicity mitigation in language models}.
\newblock \emph{Preprint}, arXiv:2407.12824.

\bibitem[{Sun et~al.(2025)Sun, Pickett, Nain, and Jones}]{Sun_Pickett_Nain_Jones_2025}
Qi~Sun, Marc Pickett, Aakash~Kumar Nain, and Llion Jones. 2025.
\newblock \href {https://doi.org/10.1609/aaai.v39i24.34708} {Transformer layers as painters}.
\newblock \emph{Proceedings of the AAAI Conference on Artificial Intelligence}, 39(24):25219--25227.

\bibitem[{Todd et~al.(2023)Todd, Li, Sharma, Mueller, Wallace, and Bau}]{todd2023function}
Eric Todd, Millicent~L Li, Arnab~Sen Sharma, Aaron Mueller, Byron~C Wallace, and David Bau. 2023.
\newblock Function vectors in large language models.
\newblock \emph{arXiv preprint arXiv:2310.15213}.

\bibitem[{Uppaal et~al.(2025)Uppaal, Dey, He, Zhong, and Hu}]{uppaal2025model}
Rheeya Uppaal, Apratim Dey, Yiting He, Yiqiao Zhong, and Junjie Hu. 2025.
\newblock \href {https://openreview.net/forum?id=lOi6FtIwR8} {Model editing as a robust and denoised variant of {DPO}: A case study on toxicity}.
\newblock In \emph{The Thirteenth International Conference on Learning Representations}.

\bibitem[{Vaswani et~al.(2017)Vaswani, Shazeer, Parmar, Uszkoreit, Jones, Gomez, Kaiser, and Polosukhin}]{vaswani2017attention}
Ashish Vaswani, Noam Shazeer, Niki Parmar, Jakob Uszkoreit, Llion Jones, Aidan~N Gomez, {\L}ukasz Kaiser, and Illia Polosukhin. 2017.
\newblock Attention is all you need.
\newblock \emph{Advances in neural information processing systems}, 30.

\bibitem[{Wang and Komatsuzaki(2021)}]{wang2021gpt}
Ben Wang and Aran Komatsuzaki. 2021.
\newblock Gpt-j-6b: A 6 billion parameter autoregressive language model.

\bibitem[{Wang et~al.(2024)Wang, Zhang, Xu, Xi, Deng, Yao, Zhang, Yang, Wang, and Chen}]{wang-etal-2024-detoxifying}
Mengru Wang, Ningyu Zhang, Ziwen Xu, Zekun Xi, Shumin Deng, Yunzhi Yao, Qishen Zhang, Linyi Yang, Jindong Wang, and Huajun Chen. 2024.
\newblock \href {https://doi.org/10.18653/v1/2024.acl-long.171} {Detoxifying large language models via knowledge editing}.
\newblock In \emph{Proceedings of the 62nd Annual Meeting of the Association for Computational Linguistics (Volume 1: Long Papers)}, pages 3093--3118, Bangkok, Thailand. Association for Computational Linguistics.

\bibitem[{Wei et~al.(2024)Wei, Deng, Pang, Ding, Shen, and Cheng}]{wei2024mlake}
Zihao Wei, Jingcheng Deng, Liang Pang, Hanxing Ding, Huawei Shen, and Xueqi Cheng. 2024.
\newblock Mlake: Multilingual knowledge editing benchmark for large language models.
\newblock \emph{arXiv preprint arXiv:2404.04990}.

\bibitem[{Xin et~al.(2024)Xin, Ren, Song, Shao, Zhao, Wang, Liu, Zhang, Lu, Du, Gao, Zhu, Yang, Gou, Wu, Luo, and Ruan}]{xin2024deepseekproverv15harnessingproofassistant}
Huajian Xin, Z.~Z. Ren, Junxiao Song, Zhihong Shao, Wanjia Zhao, Haocheng Wang, Bo~Liu, Liyue Zhang, Xuan Lu, Qiushi Du, Wenjun Gao, Qihao Zhu, Dejian Yang, Zhibin Gou, Z.~F. Wu, Fuli Luo, and Chong Ruan. 2024.
\newblock \href {https://arxiv.org/abs/2408.08152} {Deepseek-prover-v1.5: Harnessing proof assistant feedback for reinforcement learning and monte-carlo tree search}.
\newblock \emph{Preprint}, arXiv:2408.08152.

\bibitem[{Xu et~al.(2024)Xu, Jiang, Niu, Jia, Lin, and Poovendran}]{xu2024safedecoding}
Zhangchen Xu, Fengqing Jiang, Luyao Niu, Jinyuan Jia, Bill~Yuchen Lin, and Radha Poovendran. 2024.
\newblock Safedecoding: Defending against jailbreak attacks via safety-aware decoding.
\newblock \emph{arXiv preprint arXiv:2402.08983}.

\bibitem[{Yan et~al.(2025)Yan, Sun, Duan, Liu, Liu, Yin, Li, and Lei}]{yan2025benignimporttoxicjailbreaking}
Yu~Yan, Sheng Sun, Zenghao Duan, Teli Liu, Min Liu, Zhiyi Yin, Qi~Li, and Jiangyu Lei. 2025.
\newblock \href {https://arxiv.org/abs/2503.00038} {from benign import toxic: Jailbreaking the language model via adversarial metaphors}.
\newblock \emph{Preprint}, arXiv:2503.00038.

\bibitem[{Yang et~al.(2024)Yang, Sun, Ma, Liu, Yin, and Cheng}]{yang2024butterflyeffectmodelediting}
Wanli Yang, Fei Sun, Xinyu Ma, Xun Liu, Dawei Yin, and Xueqi Cheng. 2024.
\newblock \href {https://arxiv.org/abs/2402.09656} {The butterfly effect of model editing: Few edits can trigger large language models collapse}.
\newblock \emph{Preprint}, arXiv:2402.09656.

\bibitem[{Yao et~al.(2024)Yao, Zhang, Xi, Wang, Xu, Deng, and Chen}]{yao2024knowledge}
Yunzhi Yao, Ningyu Zhang, Zekun Xi, Mengru Wang, Ziwen Xu, Shumin Deng, and Huajun Chen. 2024.
\newblock Knowledge circuits in pretrained transformers.
\newblock \emph{arXiv preprint arXiv:2405.17969}.

\bibitem[{Yu and Ananiadou(2023)}]{yu2023neuron}
Zeping Yu and Sophia Ananiadou. 2023.
\newblock Neuron-level knowledge attribution in large language models.
\newblock \emph{arXiv preprint arXiv:2312.12141}.

\bibitem[{Zhang et~al.(2022)Zhang, Roller, Goyal, Artetxe, Chen, Chen, Dewan, Diab, Li, Lin et~al.}]{zhang2022opt}
Susan Zhang, Stephen Roller, Naman Goyal, Mikel Artetxe, Moya Chen, Shuohui Chen, Christopher Dewan, Mona Diab, Xian Li, Xi~Victoria Lin, and 1 others. 2022.
\newblock Opt: Open pre-trained transformer language models.
\newblock \emph{arXiv preprint arXiv:2205.01068}.

\bibitem[{Zhang et~al.(2023)Zhang, Cheng, Sun, Deng, and Huang}]{zhang2023instructsafety}
Zhexin Zhang, Jiale Cheng, Hao Sun, Jiawen Deng, and Minlie Huang. 2023.
\newblock Instructsafety: a unified framework for building multidimensional and explainable safety detector through instruction tuning.
\newblock In \emph{Findings of the Association for Computational Linguistics: EMNLP 2023}, pages 10421--10436.

\bibitem[{Zhao et~al.(2024)Zhao, Li, Li, Zhang, and Sun}]{zhao2024defending}
Wei Zhao, Zhe Li, Yige Li, Ye~Zhang, and Jun Sun. 2024.
\newblock Defending large language models against jailbreak attacks via layer-specific editing.
\newblock \emph{arXiv preprint arXiv:2405.18166}.

\bibitem[{Zhu et~al.(2023)Zhu, Zhang, An, Wu, Barrow, Wang, Huang, Nenkova, and Sun}]{zhu2023autodan}
Sicheng Zhu, Ruiyi Zhang, Bang An, Gang Wu, Joe Barrow, Zichao Wang, Furong Huang, Ani Nenkova, and Tong Sun. 2023.
\newblock Autodan: interpretable gradient-based adversarial attacks on large language models.
\newblock \emph{arXiv preprint arXiv:2310.15140}.

\bibitem[{Zou et~al.(2023)Zou, Wang, Carlini, Nasr, Kolter, and Fredrikson}]{zou2023universal}
Andy Zou, Zifan Wang, Nicholas Carlini, Milad Nasr, J~Zico Kolter, and Matt Fredrikson. 2023.
\newblock Universal and transferable adversarial attacks on aligned language models.
\newblock \emph{arXiv preprint arXiv:2307.15043}.

\end{thebibliography}

\appendix

\section{Related Works}

\subsection{Reducing Toxicity in LLMs}

% 参考Whispering Experts文章的写法

Existing approaches for reducing toxicity in large language models (LLMs) can be broadly categorized into three groups.
(1) Pre-training Data Modification. 
These methods reduce toxic generation by curating or modifying the data used during model pre-training~\cite{korbak2023pretraining, keskar2019ctrl}. 
(2) Tuning-based Methods.
This line of work fine-tunes LLMs into safer variants using supervised learning or reinforcement learning from human feedback, such as Supervised Safety Fine-Tuning (SSFT)~\cite{ouyang2022training} and Direct Preference Optimization (DPO)~\cite{dpo2023}.
(3) Toxicity Detection and Filtering.
These approaches add detection mechanisms to identify and block toxic content at the input or output level during inference~\cite{zhang2023instructsafety,qin2020back,hallinan2022detoxifying}.

Above methods do not address the underlying causes of toxicity within the model, and aligned LLMs remain susceptible to adversarial prompting attacks~\cite{zou2023universal,zhu2023autodan,yan2025benignimporttoxicjailbreaking}. Consequently, recent research has shifted toward analyzing the internal mechanisms of LLMs, with the goal of understanding and localizing the regions responsible for toxic behavior~\cite{lee2024a,suau2024whisperingexpertsneuralinterventions,pan2025hidden,uppaal2025model,wang-etal-2024-detoxifying}.

\subsection{Mechanistic Interpretability}

The goal of mechanistic interpretability is to reverse-engineer model behaviors~\cite{elhage2021mathematical} by mapping functional properties, such as knowledge~\cite{meng2022locating}, linguistic features~\cite{wei2024mlake}, toxicity~\cite{wang-etal-2024-detoxifying}, even tasks\cite{todd2023function} to identifiable components within LLMs. These components include neurons~\cite{yu2023neuron,dai-etal-2022-knowledge}, 
Multi-headed Self-attention (MHSA)~\cite{leong-etal-2023-self}, 
Feed-Forward Network (FFN)~\cite{deng2024everything,duan2025related},
Transformer layer~\cite{xu2024safedecoding,zhao2024defending}, 
and circuit~\cite{yao2024knowledge,ou2025llms}.

\section{Experimental Detail}   \label{sec:Experimental Detail}

In this section, we describe the implementation details for all baseline and proposed methods.

For \textbf{DPO}, we follow the setup of~\cite{lee2024a} and train models on 2,000 pairwise toxic samples. Default hyperparameters are used with $\beta = 0.1$. For larger models, we apply LoRA~\cite{DBLP:journals/corr/abs-2106-09685} to each layer, with a rank of 64, scaling factor of 16, and dropout rate of 0.1. Training employs early stopping with a patience value of 10 based on validation loss.

For \textbf{SSFT}, we follow the DPO setup, including dataset, LoRA, and early stopping.

For \textbf{ProFS}, we follow~\cite{uppaal2025model} and train on 500 pairwise toxic samples. Two hyperparameters are tuned: the number of top-$k$ right singular vectors for constructing the toxic subspace, and the starting layer index $\ell_0$ for projection-based editing. Specifically, we set $(k=2,\ \ell_0=11)$ for GPT-2; $(k=10,\ \ell_0=11)$ for GPT-J; and $(k=10,\ \ell_0=15)$ for all other models.

For \textbf{GloSS}, we introduce three hyperparameters: the toxicity threshold $\tau$ for selecting candidate directions, the variance ratio $\eta$ for PCA-based subspace extraction, and the starting layer index $\ell_0$ for applying projection. The detailed configurations of these hyperparameters for each model are summarized in Table~\ref{tab:hyperparameter}.

\begin{table}
  \centering
  \caption{GloSS Hyperparameters. $\tau$ and $\eta$ are used to identify the global toxic subspace, while $\ell_0$ determines the layers where projection is applied.}
  \label{tab:hyperparameter}
  \begin{tabularx}{0.45\textwidth}{Xccc}
    \toprule
    \textbf{Model} & \textbf{$\tau$} & \textbf{$\eta$} & \textbf{$\ell_0$} \\
    \midrule
    GPT-2 Medium   & 1.0 & 0.8 & 13-24 \\
    GPT-J 6B       & 4.0 & 0.7 & 15-28 \\
    OPT-6.7B       & 2.0 & 0.8 & 10-32 \\
    Mistral-7B     & 1.0 & 0.7 & 15-32 \\
    \bottomrule
  \end{tabularx}
\end{table}

\end{document}